\documentclass[conference, letterpaper]{IEEEtran}
\IEEEoverridecommandlockouts
\usepackage{cite}
\usepackage{amsmath,amssymb,amsfonts}
\usepackage{algorithmic}
\usepackage{graphicx}
\usepackage{textcomp}
\usepackage{xcolor}
\usepackage{hyperref}

\usepackage{booktabs}       

\usepackage{comment}
\usepackage{amsmath,amssymb} 
\usepackage{color}

\usepackage{subcaption}
\usepackage{multirow}
\usepackage{makecell}

\usepackage[export]{adjustbox}
\usepackage{cellspace}  
\usepackage{enumitem}

\usepackage{tikz}
\usepackage{textcomp}

\newcommand\copyrighttext{%
	\footnotesize \textcopyright 2021 IEEE. Personal use of this material is permitted. Permission from IEEE must be
	obtained for all other uses, in any current or future media, including
	reprinting/republishing this material for advertising or promotional purposes, creating new
	collective works, for resale or redistribution to servers or lists, or reuse of any copyrighted
	component of this work in other works.}
\newcommand\copyrightnotice{%
	\begin{tikzpicture}[remember picture,overlay]
	\node[anchor=south,yshift=10pt] at (current page.south) {\fbox{\parbox{\dimexpr\textwidth-\fboxsep-\fboxrule\relax}{\copyrighttext}}};
	\end{tikzpicture}%
}

\def\BibTeX{{\rm B\kern-.05em{\sc i\kern-.025em b}\kern-.08em
    T\kern-.1667em\lower.7ex\hbox{E}\kern-.125emX}}
\begin{document}

\title{Comprehensible Convolutional Neural Networks via Guided Concept Learning\\}

\author{\IEEEauthorblockN{Sandareka Wickramanayake}
\IEEEauthorblockA{\textit{School of Computing} \\
\textit{National University of Singapore}\\
sandaw@comp.nus.edu.sg}
\and
\IEEEauthorblockN{Wynne Hsu}
\IEEEauthorblockA{\textit{School of Computing} \\
	\textit{National University of Singapore}\\
	whsu@comp.nus.edu.sg}
\and
\IEEEauthorblockN{Mong Li Lee}
\IEEEauthorblockA{\textit{School of Computing} \\
	\textit{National University of Singapore}\\
	leeml@comp.nus.edu.sg}
}

\maketitle
\copyrightnotice
\begin{abstract}
	Learning concepts that are consistent with human perception is important for Deep Neural Networks to win end-user trust. Post-hoc interpretation methods lack transparency in the feature representations learned by the models. 
This work proposes a guided learning approach with an additional concept layer in a CNN-based architecture to learn the associations between visual features and word phrases.  
We design an objective function that optimizes both prediction accuracy and semantics of the learned feature representations. 
Experiment results demonstrate that the proposed model can learn concepts that are consistent with human perception and their corresponding contributions to the model decision without compromising accuracy. Further, these learned concepts are transferable to new classes of objects that have similar concepts.

\end{abstract}


\section{Introduction}
\begin{figure}[t!]
	\centering 
	\includegraphics[width=0.2\linewidth]{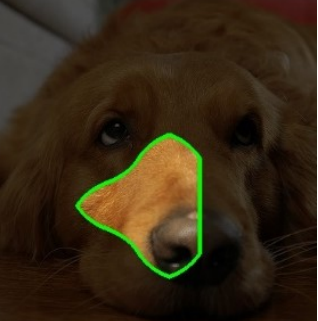} \qquad \qquad  \qquad
	\includegraphics[width=0.2\linewidth]{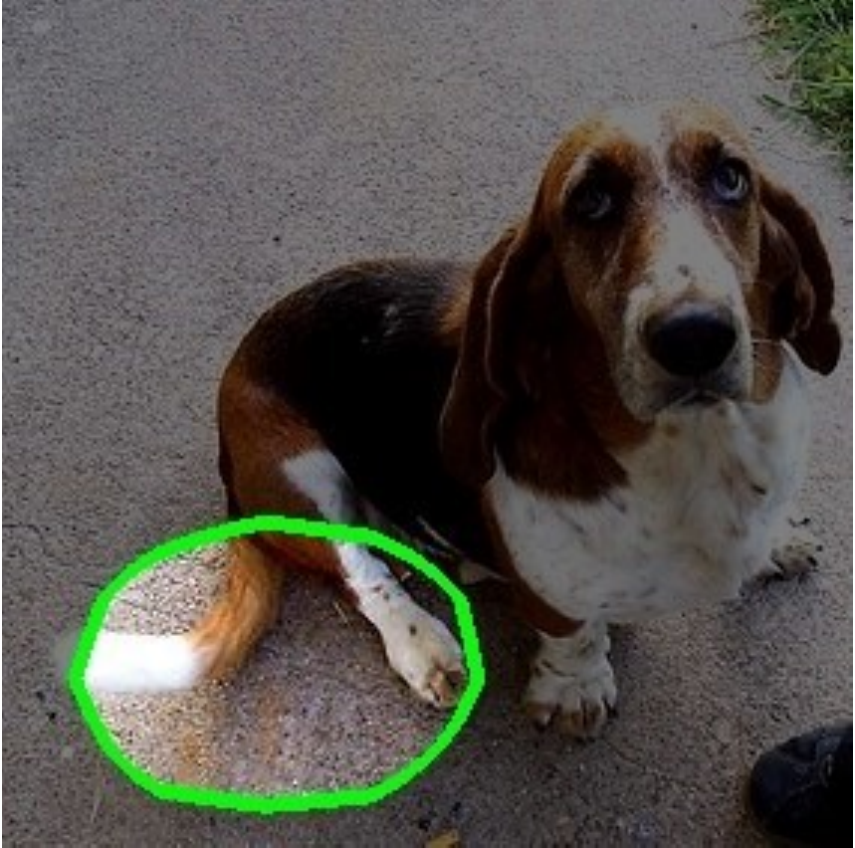} \\
	\qquad	(a) Partial region \qquad   \qquad  (b) Overlapping regions.
	\caption{Features learned by existing interpretable CNNs. }
	\label{features}	
\end{figure}

Convolutional Neural Networks (CNNs) have exhibited superior accuracy in many object detection/recognition tasks and are widely used as feature extractors in image captioning ~\cite{Vinyals2015,Xu2015} and visual question answering ~\cite{xu2016ask}. Despite such remarkable performance, CNN remains  a black box hindering user trust in AI systems~\cite{zhang2018interpretable}. 
Post-hoc interpretation methods for CNN 
~\cite{olah2018building,Hendricks2016,koh2017understanding,Lundberg2016,park2018multimodal} 
require separate modeling efforts and may not give the actual reasons behind model decisions~\cite{Sandareka2019}. 
Attempts to develop interpretable CNNs include enforcing feature disentanglement ~\cite{zhang2018interpretable} and learning a set of prototypes  ~\cite{chen2019looks,li2018deep,melis2018towards}. However, the features learned by these CNNs are not consistent with human perception and may correspond to only a partial region of a concept such as "nose" (see Figure~\ref{features}(a)), or the 
overlapping regions of two or more concepts ("leg" and "tail") as shown in Figure~\ref{features}(b).

\begin{figure}[t!]
	\centering
	\includegraphics[width=0.9\linewidth]{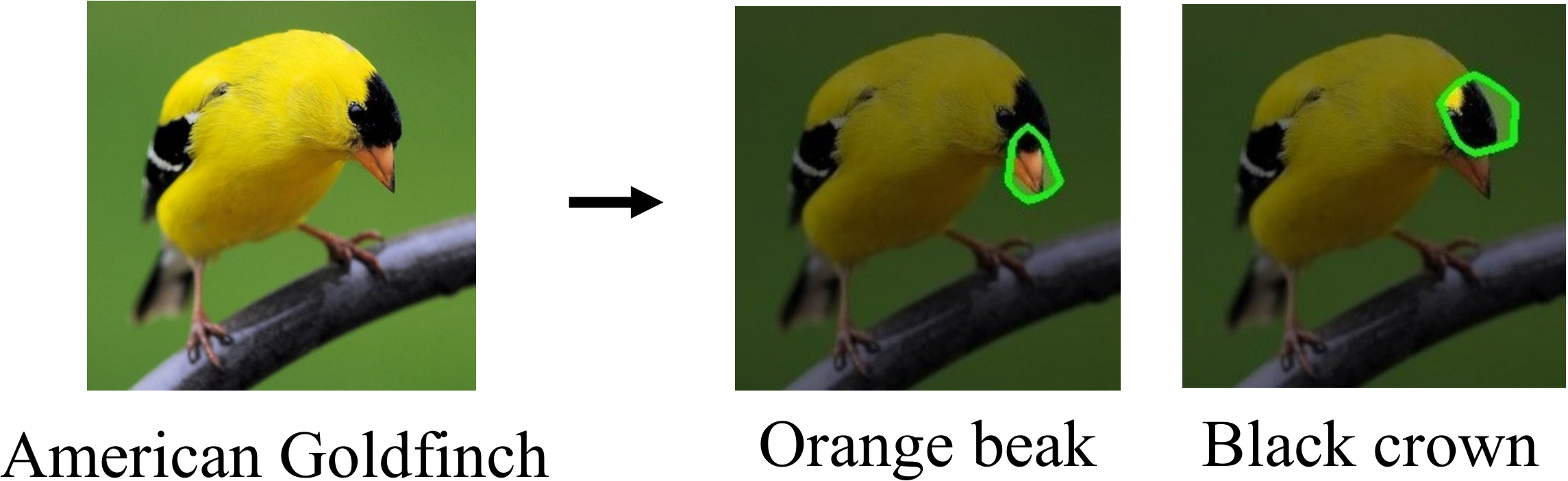} 	
	\caption{Concepts "orange beak" and "black crown" learned and their respective contributions of 0.9 and 0.1 to the class prediction "American Goldfinch".}
	\vspace{-0.025in}
	\label{example}	
\end{figure}

Ideally, the features learned by convolutional filters (conv-filters) in a CNN should correspond to concepts that are consistent with human perception and can be described in word phrases; in other words, 
we can tag  the localized image region corresponding to a learned feature with a word phrase consistent with human perception of that region.
If these concepts can be further expressed as a linear combination of their contributions to the model's decision, then we say that the CNN is comprehensible. 
For example, a comprehensible CNN would explain that it classifies the 
bird in Figure~\ref{example} to be an American Goldfinch because of its features "orange beak" and "black crown" with their respective contributions of 0.9 and 0.1 to this decision.

\begin{figure*}[t!]
	\center 
	\includegraphics[width=0.8\linewidth]{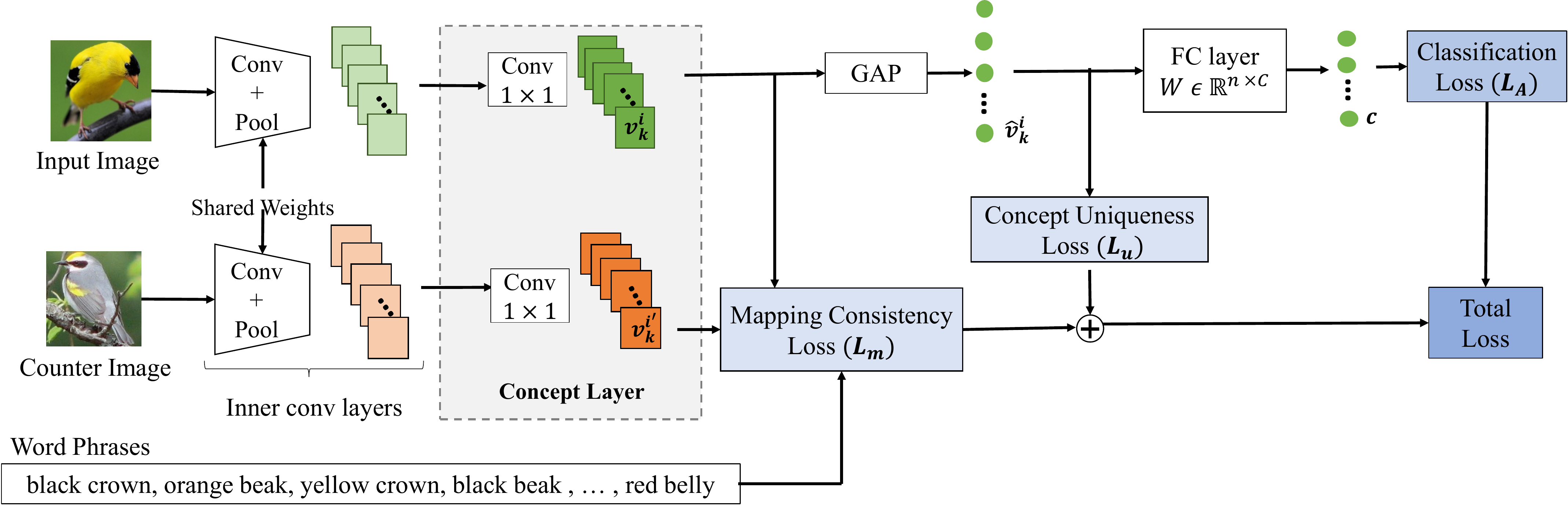}
	\caption{Overview of proposed comprehensible CNN.}
	\label{fig:overview}
\end{figure*}

In this work, we   introduce  an additional concept layer into a CNN-based architecture to guide the learning of the associations between visual features and word phrases extracted from image descriptions.
For comprehensibility, we design a new 
training objective function that  considers \emph{concept uniqueness} 
where each  learned concept corresponds to only one  word phrase, and 
\emph{ mapping consistency} which  aims to preserve the distance between the learned concept and its corresponding word phrase in a joint embedding space. 
The new training objective together with classification accuracy loss ensure that the proposed model is trained to make accurate and comprehensible decisions.
Our proposed approach employs a global pooling layer to 
express its classification decision as a weighted sum of the learned concepts. With this, we can explain the model decision in terms of  word phrases and their corresponding contributions, without the need for post-hoc processing.

Experiments on multiple datasets as well as a user study
indicate that our approach can effectively learn concepts that are comprehensible. Further, we show that these concepts are transferable and can be applied to objects of new classes with similar concepts.


\section{Related Work}

Research on opening black-box CNN can be broadly categorized into post-hoc methods and interpretable CNNs. Post-hoc methods to interpret CNN features  include visualizations \cite{zeiler2014}, activation maximization \cite{olah2018building} and quantifying interpretability of filters  \cite{bau2017,kim2017tcav}. These methods do not explain  what features contributed to a CNN decision, and   their interpretations may not be readily understood by the user. Other post-hoc methods try to explain CNN decisions via visualizations \cite{Bach2015,selvaraju2017grad,zhou2018interpretable}, super-pixel segments \cite{ghorbani2019towards,yeh2019concept} or linguistic explanations \cite{Hendricks2016,park2018multimodal}. 
However, these explanations may not be descriptive or may not capture the actual reasons behind model decisions.
The FLEX framework   generates descriptive explanations that reflect  model decisions  \cite{Sandareka2019}.  These  post-hoc interpretations require separate modeling efforts. 



For interpretable CNN, the works in \cite{chen2019looks, li2018deep}  propose to train CNN to make decisions based on a set of automatically learned prototypes. Both works explain model decisions by presenting a decision-relevant set of prototypes. However, such prototypes may contain multiple object parts and it is not clear which part is responsible for the model's decision. 
\cite{zhou2016learning} uses global average pooling to identify discriminative regions in an image and show that it can be used for object localization. Given an explanation method, the model proposed in\cite{Pillai2021} learns to produce consistent interpretations. 
\cite{zhang2018interpretable} encourages filters in the higher level convolutional layers to be activated only for a single object part. However, the learned features may not correspond to human understandable concepts, and post-hoc knowledge distilling techniques such as \cite{chen2019explaining} and \cite{zhang2019interpreting}  are needed to make sense of these features as well as their contributions  to the decision. 

\section{Proposed Comprehensible CNN }
\label{sec:ccnn}

Figure~\ref{fig:overview} shows an overview of the proposed comprehensible CNN.
We add an additional concept layer  that takes the outputs of the last convolutional layer in a standard CNN architecture such as VGG \cite{simonyan2014very}, ResNet \cite{he2016deep} or DenseNet \cite{huang2017densely}
and passes them through 1x1 convolutional filters to obtain the visual features. These visual features are then mapped to word phrases guiding the filters in the concept layer to learn concepts that are consistent with human perception.


%
Given an image and a description  of its content,
we extract a set of word phrases. 
Each image belongs to some class $c$, $c \in \{1,2,...,C\}$, and each word phrase consists of a noun and its associated adjectives depicting some concept in the image. 
Let $P$ be the set of word phrases extracted from the descriptions of images in a training dataset, and 
$P_c$ be the set of word phrases for the images in the class $c$. 
We treat each $P_c$ as a single document, and compute the tf-idf~\cite{salton1986introduction} for each word phrase $p \in P$ in the corpus of documents $\{P_1, P_2, \cdots, P_C\}$.
The final set of word phrases $P^*$ is given by $\bigcup {P^*_c}$ for $c = \{1,2,...,C\}$ where 
$P^*_c$ is the set of top phrases with the highest scores.

For each image, the filters in the concept layer aim to associate the visual features learned to some word phrases in $P^*$.
Let $\mathcal{V}^i = \{v^i_1,...,v^i_n\}$ be the set of  visual features extracted by the concept layer.
Let $g:\mathcal{V}^i \rightarrow P^*$ denotes the mapping such that  
for each $v$ in $\mathcal{V}^i$, we have
$g(v)=p$, for some word phrase $p \in P^*$.
To achieve this, we design an objective function that maximizes concept uniqueness  and mapping consistency.
Concept uniqueness  encourages one-to-one mapping whereby each concept learned  corresponds to only one  word phrase, while
mapping consistency aims to preserve the distance between a visual feature and its corresponding word phrase in a joint embedding space. 

We further improve this mapping consistency by 
comparing the differences between the input image that has a particular concept, say the $k^{th}$ concept, with a counter-image that does not have this concept.
This allows us to penalize those cases where the word phrase $p_k$ in the joint embedding space is closer to  the visual feature of the counter-image than that of the input image. 
Our proposed comprehensible CNN is optimized for classification accuracy, concept uniqueness and mapping consistency. We use cross-entropy loss $\boldsymbol{L_{A}}$ as the classification loss. The final training objective function is given by:
\begin{align*}
\boldsymbol{L} &=  \boldsymbol{L_{A}} + \lambda (  \boldsymbol{L_{u}} + \boldsymbol{L_{m}})
\end{align*}
where  $\lambda \in [0,1]$ is a regularizer,  $\boldsymbol{L_u}$ is the
concept uniqueness loss  and $\boldsymbol{L_m}$ is the mapping consistency loss. 


%

\subsection{Concept Uniqueness Loss}

Interpretability is enhanced when
each filter in the concept layer is able to learn the visual feature corresponding to only one word phrase in $P^*$, that is, we want a filter
$k$ to be activated only when the concept described by $p_k \in P^*$ occurs in the input image $i$.
The output of a filter $k$ in the concept layer is the visual feature $v^i_k$, which is then passed to a Global Average Pooling (GAP) layer to obtain $\hat v^i_k$.
We normalize $\hat v^i_k$ by subtracting the mean activation of the GAP layer to get $\tilde v^i_k$ as zero-mean normalization has been shown to increase the resilience to the scaling factor of the objects in images. 

We create an indicator vector $\boldsymbol{Z^i}$ = $(z^i_1,z^i_2,..,z^i_n)$ for the image $i$ 
where $n$ = $|P^*|$,
$z^i_k =1$ if the $k^{th}$ word phrase $p_k \in P^*_c$ is present in the image description, and $z^i_k=0$ otherwise. 
Then 
we compare $\tilde v^i_k$ with $z^i_k$ in the indicator vector $\boldsymbol{Z^i}$. 
The \emph{concept uniqueness loss} $\boldsymbol{L_{u}}$ is given by
\begin{align*}
\boldsymbol{L_{u}} &= \sum_k -(z^i_k \log(y^i_k)+(1-z^i_k)\log(1-y^i_k))
\end{align*}
where $y^i_k = \frac{1}{1+e^{-\tilde v^i_k}}$.
By minimizing this loss function,  the visual features extracted by filter $k$ is guided to uniquely represent the concept corresponding to $p_k$ because $\tilde v^i_k$ is large only when $p_k$ is present.



\subsection{Mapping Consistency Loss}\label{sec:semantic}

An image often contains multiple concepts, and a subset of these concepts is often activated together, especially for images from the same class. As a result, it is unclear which filter's output  corresponds to which specific word phrase.
For example, the bird American Goldfinch is characterized by having "orange beak" and "black crown". During training, the same two filters in the concept layer are always activated whenever they see an image of American Goldfinch. This makes it hard to 
know  which filter's output captures the concept "orange beak" or  "black crown".

We address this issue using the \emph{mapping consistency loss} $\boldsymbol{L_{m}}$ which has two components, \textit{semantic loss} $\boldsymbol{L_{s}}$  and  \textit{counter loss} $\boldsymbol{L_{c}}$, that is, 
\begin{align*}
\boldsymbol{L_{m}} &= \boldsymbol{L_{s}} + \boldsymbol{L_{c}}
\end{align*}

\begin{figure}[t!]
	\center 
	\includegraphics[width=0.7\linewidth]{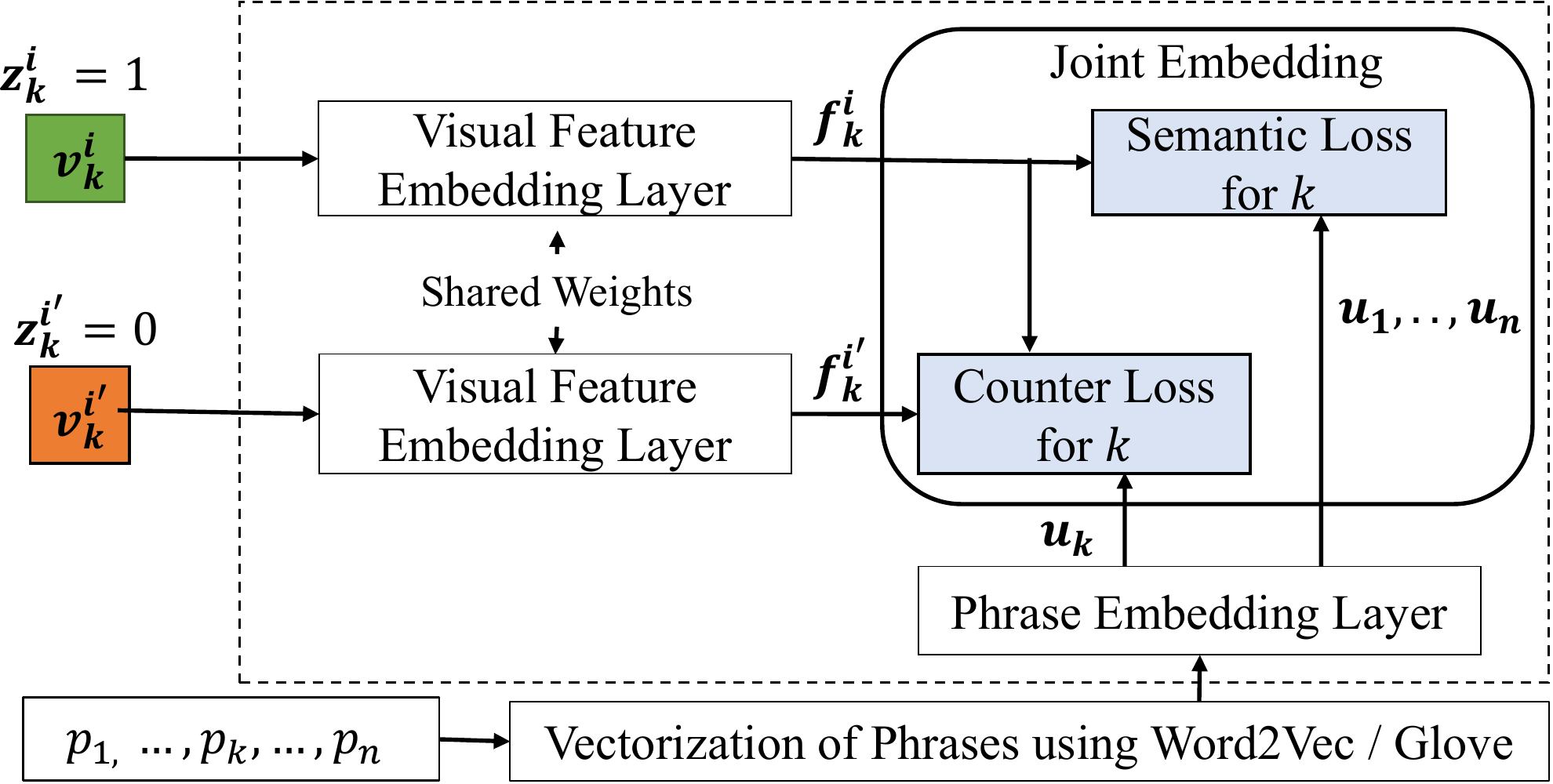}
	\caption{Mapping of visual features $v^i_k$, $v^{i'}_k$ and word phrase $p_k$ in the joint embedding space.}
	\label{fig:mapping_consistency}
\end{figure}

We map the visual features and their word phrases to a joint embedding space such that  the similarities between corresponding pairs of visual features and word phrases are preserved (see Figure \ref{fig:mapping_consistency}).  We use two embedding layers to project the visual feature $v^i_k$ and the vector representation of the corresponding word phrase
$p_k$ to a joint embedding space. 
The visual feature embedding layer embeds each  $v^i_k$  after being flattened 
and outputs $f^i_k$.

We compute the vector representation of each word phrase $p_k$  as the summation of embeddings of all the words  in $p_k$. The phrase embedding layer embeds the vector representation into the joint space and outputs $u_k$.
The similarity between  the pair of visual feature and its corresponding word phrase should be higher than that between the visual feature and any other word phrase.
We capture this constraint as the triplet loss in our objective function  $\boldsymbol{L_{s}}$ as follows: 
\begin{align*}
\boldsymbol{L_{s}} &= \sum_k z^i_k \sum^n_{k' \neq k} \max\left(0, \frac{f^i_k.u_{k'}}
{\Vert f^i_k\Vert \Vert u_{k'}\Vert} - \frac{ f^i_k.u_{k}}
{\Vert f^i_k\Vert \Vert u_{k}\Vert} + \alpha\right)
\end{align*}
where  $\alpha$ is a hyperparameter, $0 \leq \alpha \leq 1$. $\alpha$ is the margin between tuples with the correct association and tuples with the incorrect association.

\begin{table*}[tbp]
	\caption{Concepts contributed to the comprehensible CNN's classifications of sample test images.}
	\small
	\centering
	\begin{tabular}{Sc p{0.15\textwidth}p{0.25\textwidth}p{0.4\textwidth}}
		\hline
		Image & Predicted class & Concept and contribution & Ground truth description \\ \hline
		\includegraphics[valign=c,width=0.045\linewidth]{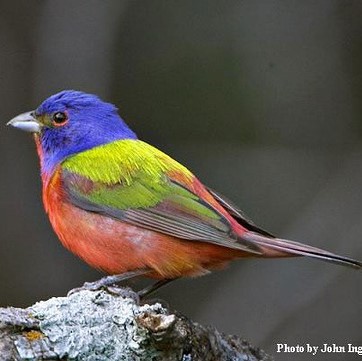} &
		Painted Bunting & 
		Green and yellow wing (0.61), \newline Blue head (0.36)  &
		A colorful bird with a blue head, light orange chest, and yellow and green wings.\\
		\hline
		\includegraphics[valign=c,width=0.045\linewidth]{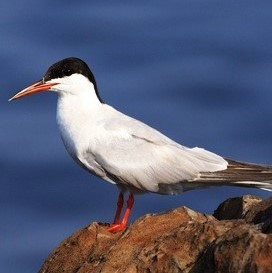} &
		Common Tern & 
		Orange beak (0.87),\newline Black head (0.06)  &
		This is a white bird with a black head and orange feet and beak. \\
		\hline
		\includegraphics[valign=c,width=0.045\linewidth]{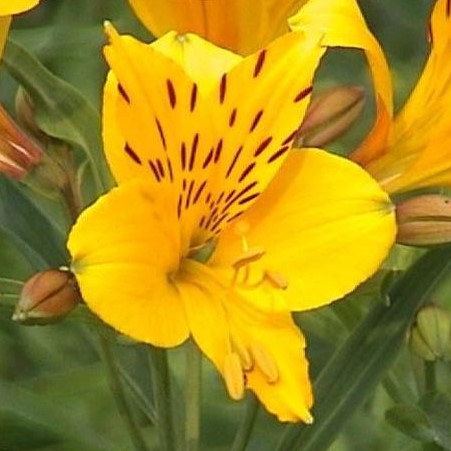} &
		Peruvian Lily& 
		Dark lines (0.94), \newline Yellow petals (0.03) & 
		This flower has very bright yellow petals, two of which have small dark red lines on them.\\
		\hline
		\includegraphics[valign=c,width=0.045\linewidth]{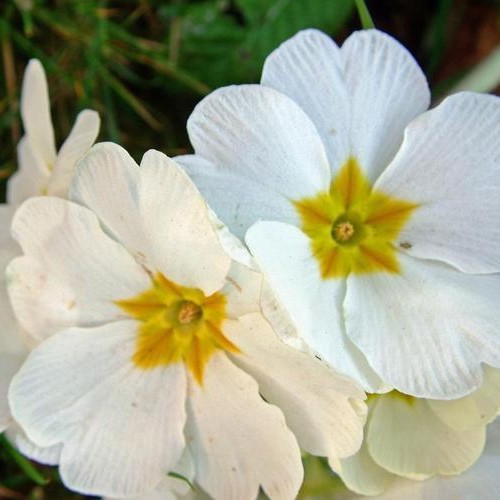} & 
		Primula & 
		Heart-shaped petals (0.98), \newline White petals (0.01)&
		The petals are heart-shaped and primarily white in color with yellow at the center and the stamen is  yellow.\\
		\hline		
	\end{tabular}
	\label{tab:case} 
\end{table*}

For a filter to isolate some concept, we compare the differences between an image $i$ that has a concept $k$ with a counter-image $i'$ that does not have this concept. In other words, the image $i'$ is a counter-image of $i$ for concept $k$ if  ${z}^i_k = 1$ and ${z}^{i'}_k = 0$. In our implementation, as a means to efficiently find a counter-image, we create a copy of the training batch and shuffle the images in the copy before pairing the  $i^{th}$ image in the original training batch with the $i^{th}$ image in the shuffled batch. For each concept $k$ that is in image $i$ but not in image $i'$ (i.e. ${z}^i_k -{z}^{i'}_k > 0$), the word phrase $u_k$ must be more similar to the visual embedding $f_{k}$  of image $i$ compared to  the visual embedding $f'_{k}$ of counter-image $i'$. This is achieved by defining the \textit{counter loss} $\boldsymbol{L_{c}}$ as
\begin{align*}
\boldsymbol{L_{c}} &= \sum_{ {z}^i_k -{z}^{i'}_k > 0}  \max\left(0,  \frac{f^{i'}_k.u_{k}}
{\Vert f^{i'}_k \Vert \Vert u_{k}\Vert} - \frac{f^i_k.u_{k}}
{\Vert f^i_k \Vert \Vert u_{k}\Vert} + \beta \right)
\end{align*}
where  $\beta$ is a hyper-parameter similar to $\alpha$, $0 \leq \beta \leq1$.

\subsection{Contribution to Classification Decision}

For comprehensibility, it is important to express the classification decision as a list of concepts that have contributed to the decision and their corresponding weights. This is achieved by utilizing 
a GAP layer to reduce the dimensionality of the outputs from the concept layer before feeding to a fully connected (FC) layer for classification (see Figure~\ref{fig:overview}).
Let $W \in \mathcal{R}^{n \times C}$ be the weight matrix of the FC layer.
We compute the contribution of concept $k$ to the decision $c$ as $\hat{v}_k \times \bar{w}_{k,c} \in W$.
After computing the contributions of all the filters in the concept layer, we apply softmax function to obtain the percentage contributions. 

Table~\ref{tab:case} shows the concepts that have contributed to the classification decisions  for  sample test images. The concepts that have contributed to the model decision  is expressed in terms of word phrases corresponding to the top-2 contributors and their respective percentage contributions. Note that these explanations directly come from the classifier, comprehensible CNN, itself rather than using an external explainer as in \cite{Sandareka2019,Hendricks2016,hendricks2018grounding}. Further, explanations from our model are more informative as they indicate the contribution of each concept towards the model decision, which is absent in existing post-hoc linguistic explanations.

\section{Experimental Study }

We conduct experiments to evaluate the accuracy and comprehensibility of our approach on the following  datasets:
\begin{itemize}	
	\item \textbf{CUB}~\cite{wah2011}.
	This UCSD Birds dataset has 11,788 bird images belonging to 200 classes.  
	Each image has ten sentences describing the bird ~\cite{Reed2016}. CUB also contains the centroid of 15 key body parts. 
	
	\item \textbf{CUB-Families}~\cite{Chen2018fine}.
	This dataset allocates 200 species in  CUB  into 37 families.
	We follow the same train/test split as in~\cite{wah2011} for evaluations. 
	
	\item
	\textbf{Flowers}~\cite{nilsback2008automated}. This dataset has 102 flower types with 2,040 training and 6,149 test images. Each image has ten sentences describing  flower species~\cite{Reed2016}. 
	
	\item
	\textbf{VOC-Part}~\cite{chen2014detect}. This is a subset of the Pascal VOC 2010 dataset containing 6 classes of animals. Each image is annotated with body parts such as head, leg, torso, etc. We also collect descriptions for the body parts of 6 animal categories from  Wikipedia. 
	
\end{itemize}
We use a rule-based concept chunker to parse the text descriptions. The concept chunker tokenizes the description and uses a POS tagger to identify nouns and adjectives. A word phrase is obtained by associating  a noun with its respective adjectives.
We select the top 20 phrases based on tf-idf scores as the  concepts for each class and obtain 398, 344, 296, and 52 phrases for CUB, CUB-Fam, Flowers, and VOC-Part respectively. We implement our Comprehensible CNN (CCNN)  on VGG16~\cite{simonyan2014very}, ResNet101~\cite{he2016deep} and DenseNet161~\cite{huang2017densely}.  Unless stated otherwise, we use VGG16~\cite{simonyan2014very} in our experiments. We set  $\lambda=0.4, \alpha=1 $, $\beta=0.5$.  The joint embedding space dimension is set to 24 for CUB and CUB Families, and 16 and 12  for Flowers and VOC-Part respectively. We train the two embedding layers along with other layers of CCNN together.
The supplementary material and code are available at \textbf{\textit{https://github.com/sandareka/CCNN}}.


\subsection{Ablation Study}
\label{ablation}

We first examine the effect of the loss functions $\boldsymbol{L_A}$, $\boldsymbol{L_u}$ and  $\boldsymbol{L_m}$ using CUB and CUB-Fam  with their ground truth part annotations where the  centroid of each part is marked.

\begin{table}[tbp]
	\caption{Results of ablation study. Mean and standard deviation over five runs.}
	\small
	\centering
	(a) CUB dataset\\
	\smallskip
	\begin{tabular}{lcc}\hline
		Model &  Classification Acc. &  Association Acc.  \\ \hline
		$\boldsymbol{L_{A}}$ &  73.3$\pm$0.2  & -   \\
		$\boldsymbol{L_{A}}$ + $\boldsymbol{L_{u}}$ & 77.2$\pm$0.1 & 78.8$\pm$0.7   \\
		$\boldsymbol{L_{A}}$ + $\boldsymbol{L_m}$ & 74.9$\pm$0.5 & 61.9$\pm$1.1  \\
		$\boldsymbol{L_{A}}$ + $\boldsymbol{L_{u}}$ + $\boldsymbol{L_m}$ & \textbf{80.1$\pm$0.5} &  \textbf{84.1$\pm$0.6} \\		
		\hline
	\end{tabular}\\ \smallskip \smallskip
	(b) CUB-Fam dataset\\ \smallskip
	\begin{tabular}{lcc}\hline
		Model &  Classification Acc. &  Association Acc.  \\ \hline
		$\boldsymbol{L_{A}}$  &  88.2$\pm$0.1  & - \\
		$\boldsymbol{L_{A}}$ + $\boldsymbol{L_{u}}$ & 89.2$\pm$0.2 & 78.6$\pm$1.0  \\
		$\boldsymbol{L_{A}}$ + $\boldsymbol{L_m}$ & 88.6$\pm$0.1 & 72.6$\pm$1.5  \\
		$\boldsymbol{L_{A}}$ + $\boldsymbol{L_{u}}$ + $\boldsymbol{L_m}$ &  \textbf{90.6$\pm$0.3} & \textbf{81.7$\pm$0.7} \\		
		\hline
	\end{tabular}
	\label{tab:tbleablation}
\end{table}

Besides classification accuracy, we also define a metric called 
association accuracy to measure the percentage of filters in the concept layer that are correctly associated to the corresponding word phrases. We say that an association is correct when the receptive field of a filter overlaps the ground truth annotation and the word phrase is consistent with localized image regions. For example, if the localized image regions show a 'black beak' but the associated word phrase is 'red beak', then this is a wrong association.

We select the top 10 images with the highest activations for a given filter $k$ because the maximum number of images in some classes in CUB test dataset is 10.  
We derive the receptive field of a filter using the method in \cite{bau2017}. Let $X_k$ denote the set of feature maps extracted by filter $k$. We compute the distribution of activations over all the spatial locations $[i,j]$ in $X_k$. We create a binary mask where activation values greater than 0.995 of the maximum activation are set to 1, and the rest are set to 0.
Then we use bilinear interpolation to generate the image-resolution mask, and overlay the  mask on an image to identify the receptive field of $k$. 
Let $s_k$ be the percentage of correct associations obtained by filter $k$.
The association accuracy of a model is given by $ \frac{1}{|P^*|}\sum_k s_k $.

Table~\ref{tab:tbleablation}   shows that
the best results for both classification and association accuracies are obtained when all three losses are considered.
Models that include $\boldsymbol{L_u}$ or/and  $\boldsymbol{L_m}$  have more discriminative power over  models trained  with $\boldsymbol{L_A}$ only.
This is because $\boldsymbol{L_u}$ and  $\boldsymbol{L_m}$ guide the  models to learn the visual features that are specific to each class of images.

%
%

\begin{figure}[t!]
	
	{\scriptsize ~All losses~~~ $L_m$ omitted~~~ $L_u$ omitted~~~~ ~All losses ~~~$L_m$ omitted ~~~$L_u$ omitted}	\\
	\centering
	\includegraphics[width=0.45\linewidth]{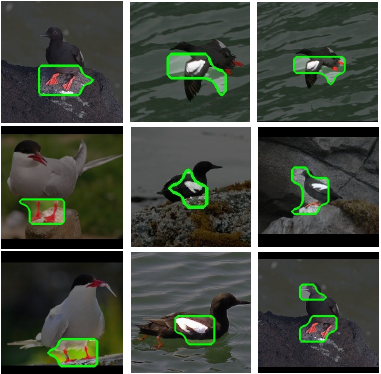}~~~~
	\includegraphics[width=0.45\linewidth]{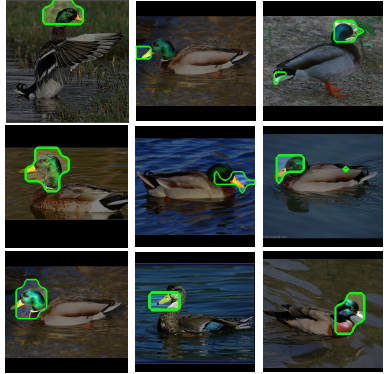} 
	\\
	(a) Filter 217 - Red leg 
	\qquad	
	(b) Filter 111 - Green head
	\caption{Visualization of sample conv-filters for CUB.}	
	\label{fig:ablation_study}
\end{figure}

Figure \ref{fig:ablation_study} shows sample filter visualizations of the ablation models  for CUB. 
Most of the training images containing "red leg" also have "white patch". Hence, the two filters corresponding to these two concepts are activated together, and in the absence of mapping consistency loss $\boldsymbol{L_m}$, the model could not differentiate whether the filter is associated to  "red leg" or "white patch". As a result, filter 217 which is trained to be activated for "red leg" is wrongly activated for "white patch". Similarly, the concepts 'green head' and 'yellow beak' often co-occur in the training images causing filter 111 to learn the wrong concept ('yellow beak' instead of 'green head') when $\boldsymbol{L_m}$ is omitted. 
When $\boldsymbol{L_u}$ is excluded, a filter may be activated for more than one concept. For example, filter 217 is activated for multiple concepts ('red leg', 'white patch' and 'black beak') while filter 111 is activated for 'green head', 'yellow beak' and 'black tail'. 

\subsection{Comparative Study}
We compare the classification accuracy of CCNN with state-of-the-art interpretable classifiers CAM~\cite{zhou2016learning},  InterpCNN~\cite{zhang2018interpretable}, ProtoPNet~\cite{chen2019looks} and CI-GC~\cite{Pillai2021}. CAM replaces fully connected layers with a GAP layer and provides saliency maps indicating image regions influenced its classification decisions. InterpCNN forces its conv-filters to be activated on the same region of an object across different images. ProtoPNet adds a new prototype layer and learns a set of prototypes for each class. CI-GC encourages the model to learn consistent interpretations using Grad-CAM~\cite{selvaraju2017grad} in addition to maximizing the log-likelihood of the correct class.

We observe that images of the same class often share similar visual characteristics and have similar descriptions. Hence, we also train a model using descriptions of a handful of images from each class and call it CCNN*. 
We apply k-means to cluster the images in each class and assume the description of the clustroid image describes images in the cluster, which may not be necessarily true.

Table~\ref{tab:comparative_acc} shows the results when $k = 5$. We see that CCNN achieves the highest accuracy, while CCNN* trained using limited and possibly incorrect image descriptions remains robust with only a slight drop in accuracy.
No results are obtained for InterpCNN on CUB and Flowers datasets as the training fails to converge due to the large number of classes. 

Table~\ref{tab:arch_comparative} compares the accuracies of interpretable classifiers when implemented using  DenseNet161 and ResNet101. We observe that CCNN consistently  achieves the highest classification accuracy. 
Table~\ref{tab:state_of_the_art} shows the classification accuracy of CCNN against
 deep learning models which are specifically designed for fine-grained classification, namely  
 	Region-CNN~\cite{huang2020interpretable},		
 TASN~\cite{zheng2019looking}, and
 MGN-CNN~\cite{zhang2019learning}. We see that CCNN has comparable accuracies with these black-box methods.  In fact, we can combine  
multiple CCNNs implemented on different architectures (e.g., CCNN-Res101, CCNN-Dense161 and CCNN-VGG16) to further improve the accuracy without compromising the interpretability.

%

\begin{table}[t!]	
	\centering
	\small
	\caption{Classification accuracy of interpretable classifiers based on VGG16.}
	 \vspace*{-0.2in}
	\begin{tabular}{p{0.20\linewidth}>{\centering}p{0.06\linewidth}>{\centering}p{0.16\linewidth}>{\centering}
			p{0.08\linewidth}>{\centering\arraybackslash}p{0.17\linewidth}} \\ \toprule
		Model & CUB &  CUB-Fam & Flowers & VOC-Part  \\ \midrule
		CAM & 70.3 &  85.8 & 92.4 & 89.6 \\			
		InterpCNN & -  & 78.4  & - & 72.4\\
		ProtoPNet & 69.8  & 86.6  & 86.9 & 84.3\\
		CI-GC & 75.3  & 89.1  & 93.3 & 90.0\\ \midrule
		CCNN & \textbf{80.1} &  \textbf{90.6}& \textbf{93.5} & \textbf{90.9}   \\ 
		CCNN* & 78.9 &  89.3 & 92.5  &  90.9  \\ 
		\bottomrule
	\end{tabular}
	\label{tab:comparative_acc}  
	 \vspace*{0.1in}	
\end{table}

\begin{table}[!t]
	\caption{Classification accuracy of interpretable classifiers using DenseNet161 and ResNet 101 architectures.}
	 \vspace*{-0.2in}
	\small
	\centering
	\begin{tabular}{lcc|cc}\\ \toprule
		\multirow{2}{*}{Model} & \multicolumn{2}{c}{CUB} & \multicolumn{2}{c}{Flowers} \\\cline{2-5}
		&  \thead{ DenseNet161} &  \thead{ ResNet101} & \thead{ DenseNet161} &  \thead{ ResNet101} \\ \hline	
		CAM & 84.2  & 85.7 & 90.1 & 96.4\\
		ProtoPNet   & 76.6  & 72.6 & 94.0 &  90.9  \\
		CI-GC   & 81.6  & 77.6 &  96.4 &  95.8  \\
		CCNN  &  \textbf{85.7} &  \textbf{86.8} & \textbf{96.8} &  \textbf{96.7}\\ 		
		\bottomrule
	\end{tabular}
 \vspace*{0.1in}
	\label{tab:arch_comparative}
\end{table}


\begin{table}[!t]	
	\centering
	\small
	\caption{Comparison of classification accuracy of CCNN with fine-grained classifiers on CUB.}
	 \vspace*{-0.2in}
	\begin{tabular}{lc} \\ \toprule
		Model & Accuracy  \\ \midrule		
		Region-CNN~\cite{huang2020interpretable} & 87.3 \\			
		TASN~\cite{zheng2019looking} & 87.9 \\
		MGN-CNN~\cite{zhang2019learning} & 89.4  \\ \midrule
		CCNN (ResNet101) & 86.8\\ 
		 Multiple CCNNs (VGG16, DenseNet161, ResNet101)  & 88.8\\ 
		\bottomrule
	\end{tabular}
	\label{tab:state_of_the_art}  	
\end{table}


Next, we compare the homogeneity of the filters in these models as is done in \cite{zhang2018interpretable}. 
Let $D$ be the set of images for which a filter $k$ is activated and
$A$  the set of ground truth part annotations.
The homogeneity of a filter $k$ is given by
\begin{align*}
\mathcal{H}(k) &= \frac{\max_{ a } ~  |I_k^a| } {|D|}
\end{align*}
where  
$I_k^a$ is the set of images  a filter $k$ is activated for some part $a \in A$.
If a filter is activated for various parts in different images, the $\mathcal{H}(k)$ value will be less than 1. Conversely, a value close to 1 implies that the filter $k$ has been activated for the same part $a$ in nearly all the images. In other words, we are confident that $k$ is highly associated with the part $a$.

Further, a filter should learn only one concept. We introduce a new metric called singleness of filters to quantify this:
\begin{align*}
\mathcal{S}(k) &= \frac{1}{|D|}\sum_{i=1}^{|D|} \frac{1}{overlap(k, A_i)}
\end{align*}
where $A_i$ is the set of part annotations for image $i$ and $overlap(k, A_i)$ is the number of annotations in $A_i$ that overlaps with the receptive field of $k$. A low value of $\mathcal{S}(k)$ indicates that the visual feature of filter $k$ covers partial regions of many concepts, leading to difficulty in understanding what has been learned. Conversely, if $\mathcal{S}(k)$ is close to 1, we are confident that the filter $k$ covers  only one concept.


We use CUB-Fam  for this experiment and evaluate the filters against four concepts: 'head', 'torso', 'legs' and 'tail'. The 'head' concept consists of the parts beak, eyes, crown, nape, forehead and throat while
the 'torso' consists of  breast, belly, back, and wings.
Since CUB-Fam provides point annotations for the body parts, we consider that filter $k$ is activated for a concept if its receptive field covers at least one of the point annotations of that concept. Otherwise, the filter is activated for the concept background. 

\begin{figure}[!t]
	\centering	
	\begin{subfigure}{0.75\linewidth}
		\centering
		\includegraphics[width=\linewidth]{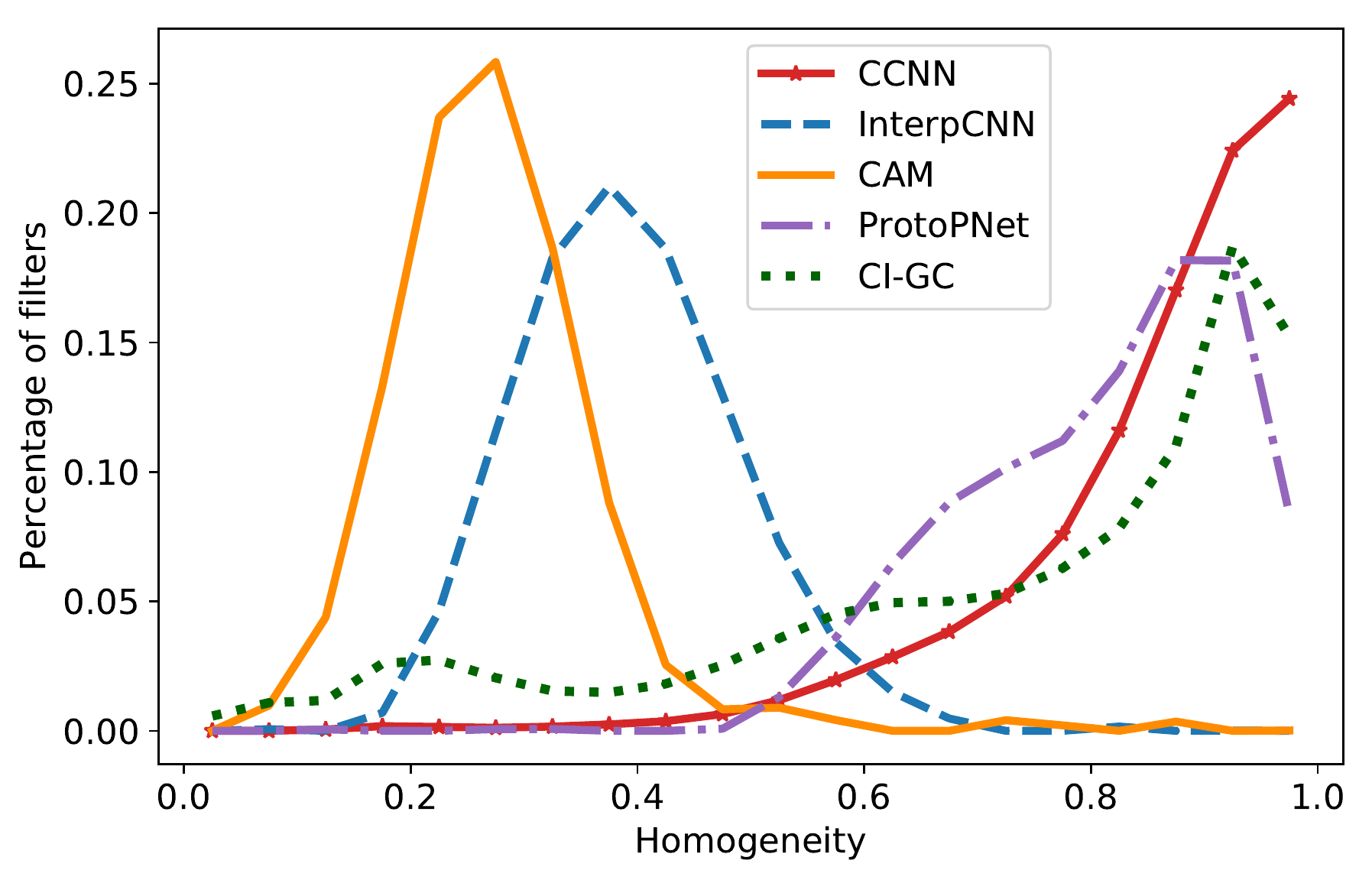}
	\end{subfigure} 
	\begin{subfigure}{0.75\linewidth}			
		\centering 
		\includegraphics[width=\linewidth]{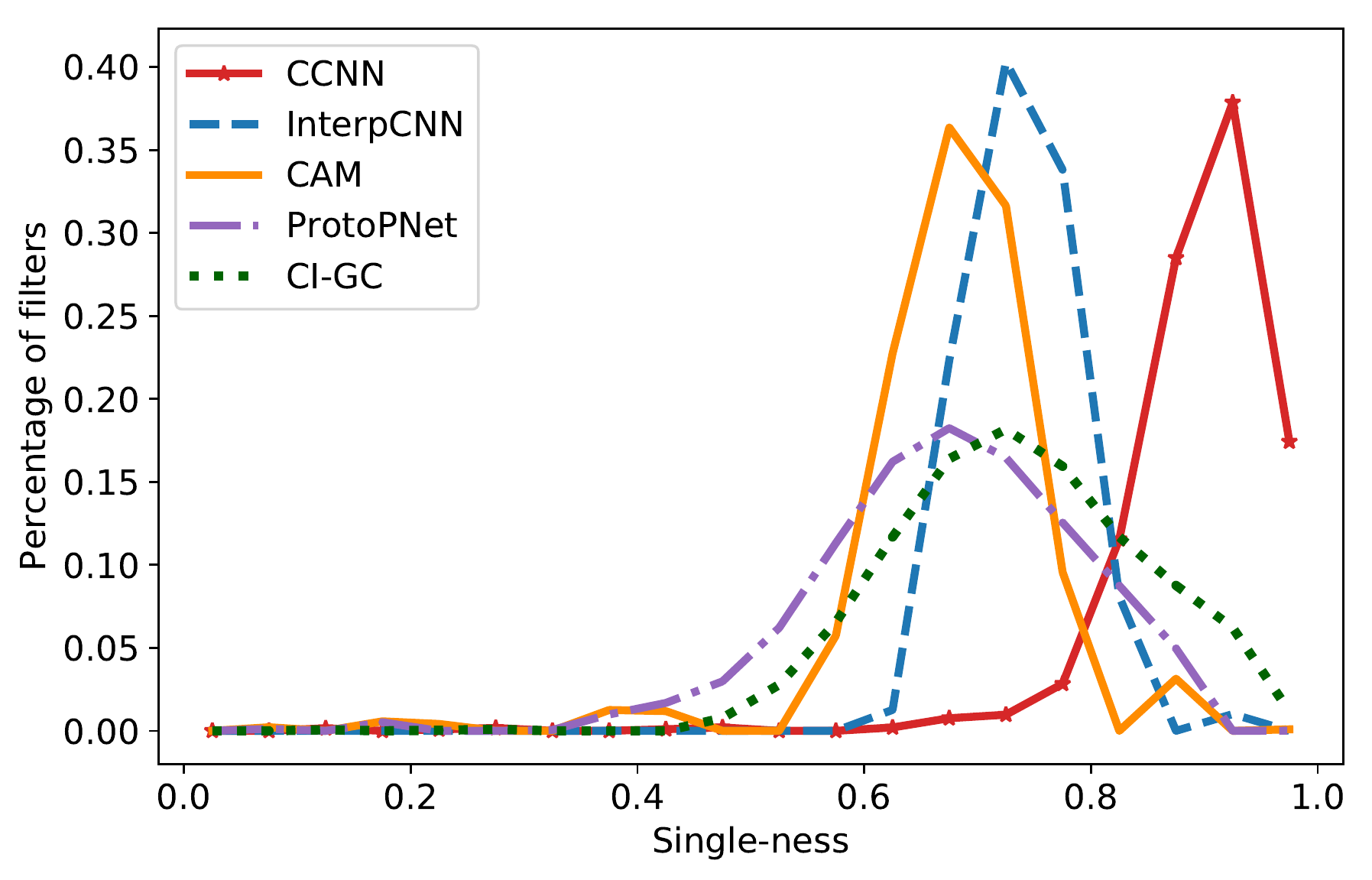}	
	\end{subfigure}	
	\caption{Distributions of $\mathcal{H}$ and $\mathcal{S}$ values of filters in CCNN, InterpCNN, CAM, CI-GC and ProtoPNet for CUB-Fam.}
	\label{fig:distribution}	
\end{figure}

Figure~\ref{fig:distribution} shows the distributions of the homogeneity and singleness  values for the filters in  the various methods. 
CCNN has a larger percentage of filters with high $\mathcal{H}$ values compared to other methods, indicating that its filters are consistently associated with the same part. 
CCNN has the highest percentage of filters whose $\mathcal{S}$ values are more than 0.9 compared to other models, which  peak around 0.70,  indicating that the filters in CCNN are mostly extracting one concept.

\begin{figure*}[!t]
	\centering	
	\includegraphics[width=0.85\linewidth]{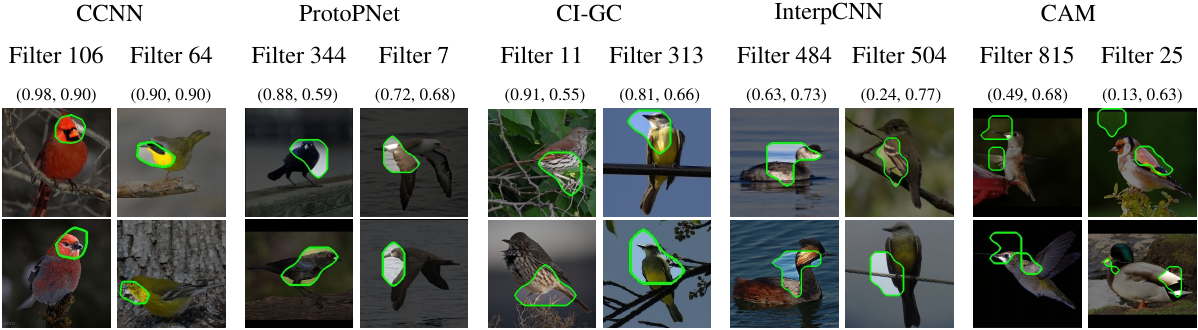}	
	\caption{Visualization of sample conv-filters for CUB-Fam and their  $\mathcal{H}$ and $\mathcal{S}$ values.}	
	\label{fig:comparison_cub}	
\end{figure*}

\begin{figure*}[!t]
	\centering
	\includegraphics[width=0.85\linewidth]{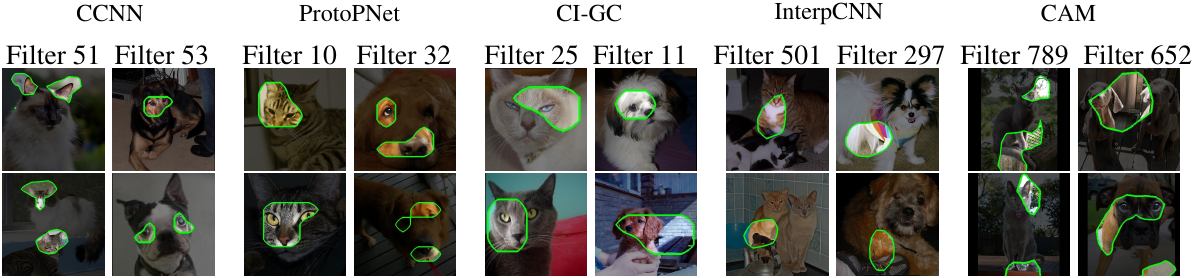}	
	\caption{Visualization of sample conv-filters in CCNN, ProtoPNet, CI-GC, InterpCNN and CAM for the VOC-Part dataset.}	
	\label{fig:comparison_voc}
\end{figure*}

\begin{figure}[!t]
	\centering
	\includegraphics[width=\linewidth]{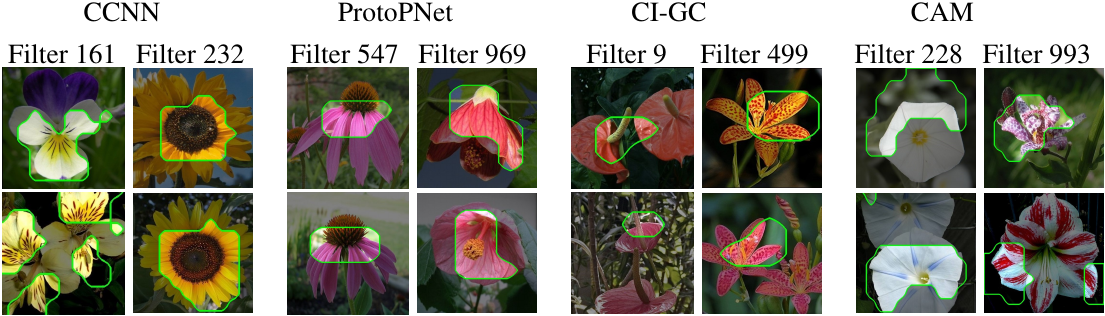}	
	\caption{Visualization of sample conv-filters in CCNN, ProtoPNet, CI-GC and CAM for Flowers dataset.} 
	\label{fig:comparison_flowers} 
\end{figure}

Figure~\ref{fig:comparison_cub} gives the visualizations of filters with different $\mathcal{S}$ and $\mathcal{H}$ values for sample images in CUB-FAM. 
We observe that filter 484 of InterpCNN partially covers the head, neck, and back of a bird with a low $\mathcal{S}$ value of 0.73.
Filter 25 in CAM has a very low $\mathcal{H}$ value of $0.13$ as it is  activated for different concepts (back of the bird in one image, and tail 
in another image). Compared to InterpCNN and CAM, filters in ProtoPNet and CI-GC have high  $\mathcal{H}$ and low $\mathcal{S}$ values as their receptive fields cover multiple object parts.
In contrast, CCNN filters are consistently activated by the same  concept  across multiple images. 
One  advantage of CCNN is that its filters learn semantically meaningful concepts even when the concept spans multiple non-contiguous regions in an image. Figures~\ref{fig:comparison_voc} and  \ref{fig:comparison_flowers} give the visualizations of the filters for concepts with multiple regions in the VOC-Part  and Flowers datasets respectively.

\begin{table*}[t!]
	\caption{Concepts transferred for CCNN-AWA and CCNN-NAB models.}
	\vspace*{-0.2in}	
	\small
	\centering
	\begin{tabular}{llll} \\ \toprule 
		{ Model} & { Class} & { Correctly transferred concepts}  & { Incorrectly transferred concepts }\\ \midrule
		\multirow{5}{*}{ CCNN-AWA}& { Deer}  & {horn, long neck, long head, pointed head} & {feathery tail, claw} \\	\cline{2-4}
		& { Fox}  &  {pointed ear, furry torso, gaskin, dog head} & {small nose, round head, wing} \\	\cline{2-4}
		& { Otter} &  {furry torso, round head, wet nose, short neck} & {feathery tail, feathery torso}  \\ \cline{2-4}
		& { Polar bear} & {furry torso, short neck, small ear, dog head} & {beak, large eye} \\ \cline{2-4}
		& { Rabbit} & {furry torso, pointed head} & {round head, feathery neck}\\ \hline
		\multirow{2}{*}{ CCNN-NAB}& { Eagle} & {large wing, yellow beak, white head, hooked beak} & {spotted side, long throat, white back} \\	\cline{2-4}
		& {  Bluebird} &{blue crown, blue head, white belly, orange breast} & {small eye, blue wingbar, blue beak} \\	
		\bottomrule
	\end{tabular}
	\label{tab:generalizability_acc}  
\end{table*}

\begin{figure}[htbp]
	\centering
	\begin{subfigure}{0.45\linewidth}
		\begin{minipage}{\linewidth}	
			\begin{subfigure}{0.45\linewidth}
				\centering\begin{scriptsize} {horn, long head, pointed head.} \\\end{scriptsize}					
			\end{subfigure}
			\begin{subfigure}{0.45\linewidth}
				\centering\begin{scriptsize} {pointed ears, long head, long neck.} \\\end{scriptsize}	
			\end{subfigure}
		\vspace{0.05in}
		\end{minipage}
		\begin{minipage}{\linewidth}
			\begin{subfigure}{0.45\linewidth}			
				\centering
				\includegraphics[width=0.9\linewidth]{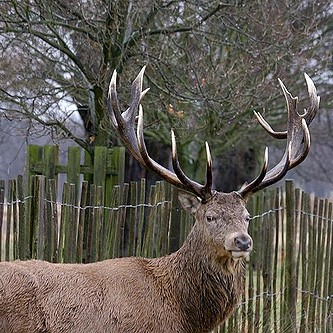}  			
			\end{subfigure}
			\begin{subfigure}{0.45\linewidth}
				\centering
				\includegraphics[width=0.9\linewidth]{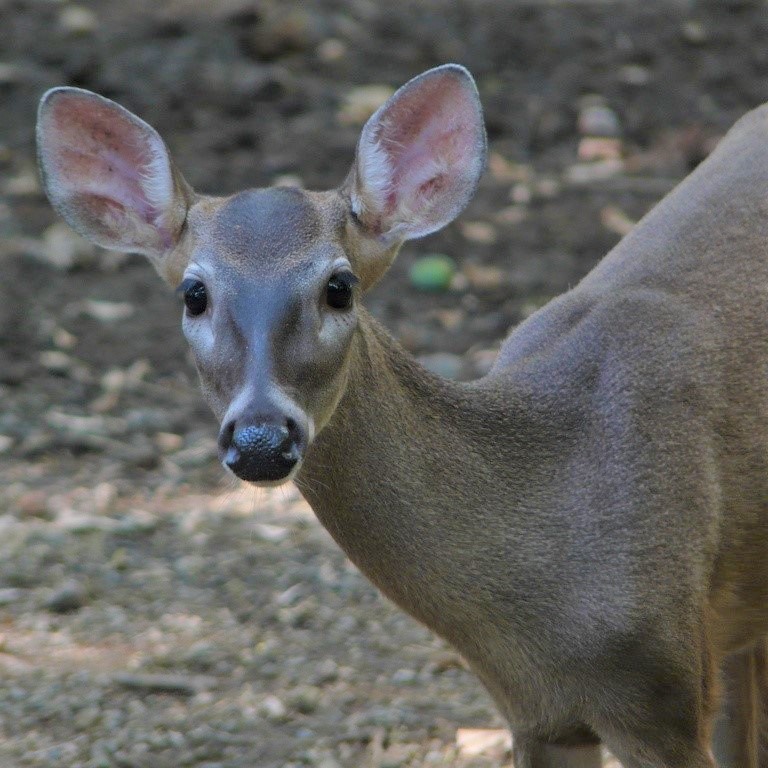} 			
			\end{subfigure}
		\end{minipage}
		
		\caption{CCNN\_AWA}
		\label{generalizability_awa}
	\end{subfigure}
	\begin{subfigure}{0.45\linewidth}
		\centering
		\begin{minipage}{\linewidth}	
			\begin{subfigure}{0.48\linewidth}
				\centering\begin{scriptsize} {yellow beak, hooked beak.} \\  \end{scriptsize}	
			\end{subfigure}
			\begin{subfigure}{0.48\linewidth}
				\centering\begin{scriptsize} {long wings, hooked beak.} \\  \end{scriptsize} 
			\end{subfigure}
		\vspace{0.001in}
		\end{minipage}
		\begin{minipage}{\linewidth}
			\begin{subfigure}{0.47\linewidth}			
				\centering
				\includegraphics[width=0.9\linewidth]{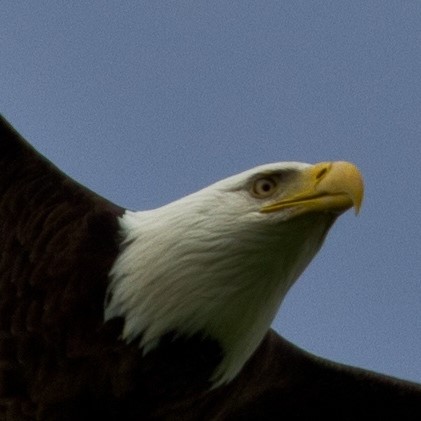} 		
			\end{subfigure}
			\begin{subfigure}{0.47\linewidth}
				\centering
				\includegraphics[width=0.9\linewidth]{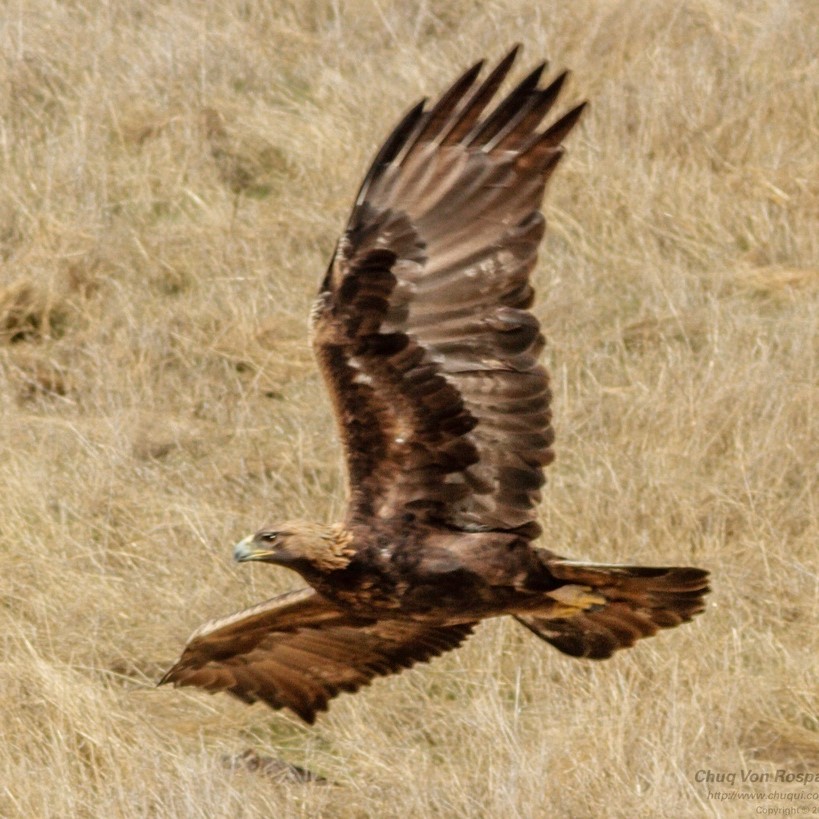}  	
			\end{subfigure}
		\end{minipage}
		\caption{ CCNN\_NAB}
		\label{generalizability_nab}
	\end{subfigure}
	\caption{Example of transferred concepts.  }
	\label{generalizability}
\end{figure}

%

\begin{figure}[htbp]
	
	\centering
	\begin{subfigure}{0.35\linewidth}			
		\centering
		\includegraphics[width=0.6\linewidth]{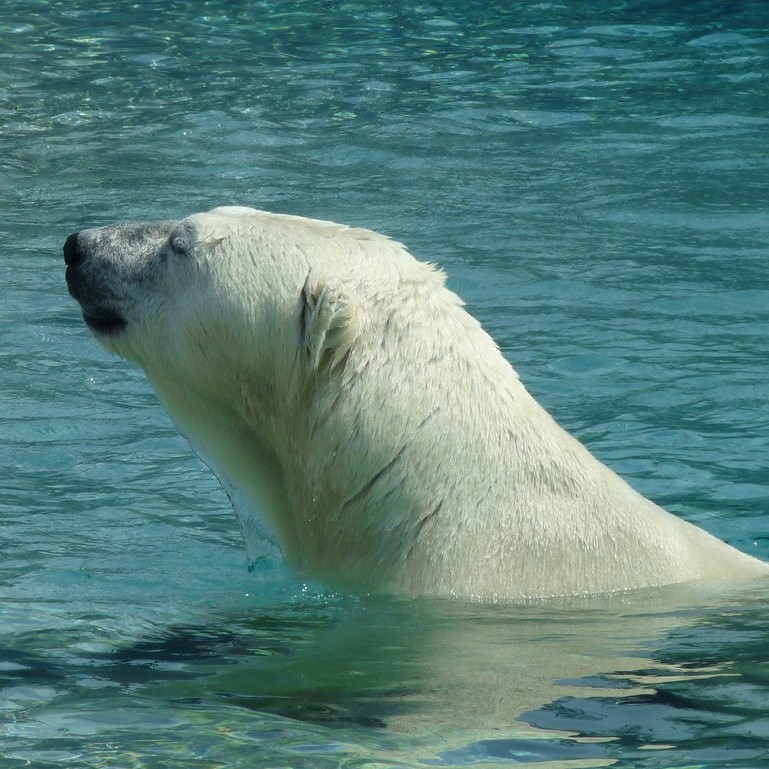}  			
	\end{subfigure} \qquad
	\begin{subfigure}{0.35\linewidth}
		\centering
		\includegraphics[width=0.6\linewidth]{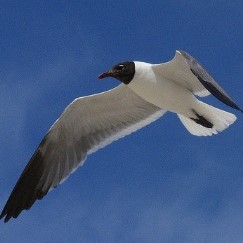} 			
	\end{subfigure}
	\caption{Polar Bear's muzzle has been mistaken as a beak due to its visual similarity with the beak.}
	\label{visual_similarity}	
\end{figure}

\subsection{Transferability of Learned Features}

In this section,  we demonstrate that the features learned by CCNN can be easily transferred to classify new objects that share similar concepts, e.g., the 'horns' concept from the class 'Cow' can be used to identify the horns of deer, buffalo, etc. 
Given a CCNN that has been pre-trained using VOC-Part, we freeze the weights of all the convolutional layers, including the concept layer and fine-tune the fully connected layer  on a new dataset consisting of five classes from the AWA dataset~\cite{awa}, namely 'Deer', 'Rabbit', 'Polar bear', 'Otter' and 'Fox'. We call this model CCNN-AWA. We also train another model, called CCNN-NAB, that uses the CCNN pre-trained on CUB and fine-tune on a new dataset having two classes of NABirds dataset~\cite{van2015building}, namely Eagle and Bluebird. 

CCNN-AWA achieves 0.96 classification accuracy while CCNN-NAB is able to classify all the test images correctly. 
We consider a concept to be transferred if its contribution to the model decision is higher than 0.05 during the classification of images of that class. A concept is said to be correctly transferred if the concept is consistent with the human perception of that class.
Table~\ref{tab:generalizability_acc} shows examples of  correctly 
and incorrectly transferred concepts for each class.

Figure~\ref{generalizability}  shows the transferred concepts that contributed to the predictions of CCNN-AWA and CCNN-NAB models for sample images. Even though no annotations were used to train these two classifiers, they are able to correctly apply the learned concepts to classify images of new classes.

We perform further analysis of concepts that have been incorrectly transferred,
and observe that in many cases, this is due to the visual similarity in the images. For example, the nose of the polar bear in Figure~\ref{visual_similarity} looks  similar to the beak of the bird because of the camera angle and the image background. This indicates that CCNN can learn visual features that are consistent with human perception, and the learned features can be transferred to new classes with similar concepts to make decisions.


\subsection{User Study on Comprehensibility}

We conduct a user study to evaluate the comprehensibility of the learned concepts by CCNN compared to CAM, InterpCNN and ProtoPNet. We randomly select 40 images from the CUB-Fam test set and for each of these images visualize the receptive field of the most activated filter of each model for that image. Each visualization indicates a single concept extracted by a model from an image. The visualizations of each image, along with a diagram indicating the regions of different body parts of the bird as in CUB~\cite{wah2011}, are shown to the users. 
Note that users do not know which visualization is generated by which model.
For each image, the order of the visualizations shown is randomized and
users are asked to  rank the visualizations  on a scale 1 to 4 with 1 indicating the most comprehensible visualization that best localize a body part.
Twenty volunteers particpated in our user study. As for the user profiles of those volunteers, 8 were female and 12 were male. The average age of a participant is 30 and standard deviation is 3.1. All participants have a bachelor’s degree. Table~\ref{tab:user_study} shows the average ranking from our study. CCNN has achieved the best average ranking indicating that users prefer the concepts learned by CCNN.

\begin{table}[t!]	
	\centering
	\small
	\caption{Comparison of comprehensibility of learned features among interpretable classifiers.}
	 \vspace*{-0.15in}
	\begin{tabular}{lc} \\ \toprule
		Model &  Human rank (1-4)  \\ \midrule		
		CAM & 3.18  \\			
		InterpCNN & 2.74\\
		ProtoPNet & 2.21  \\ 
		CCNN & 1.88 \\ 
		\bottomrule
	\end{tabular}
	\label{tab:user_study}  	
\end{table}

\section{Conclusion}
In this work, we have proposed a guided approach to enable the learning of comprehensible concepts by  associating visual features to word phrases through the addition of a concept layer in the CNN architecture. We have designed an objective function that optimizes for classification accuracy, concept uniqueness and mapping consistency.
We have introduced the homogeneity and singleness metrics to quantify how well the filters in a model  are consistently associated to a concept.
Experiments on benchmark datasets demonstrated that CCNN is able to provide the reasons for its classification decision using word phrases and a user study conducted has shown that CCNN's learned concepts are  comprehensible. Further, the learned concepts can be transferred to classify new objects with similar visual features.

\section*{Acknowledgement}
This research is supported by the National Research Foundation, Singapore under its
AI Singapore Programme (AISG Award No: AISG-GC-2019-001 and AISG-RP-2018-008). 

\bibliographystyle{ieeetr}
\bibliography{ccnn}



\end{document}